# Are Models Trained on Indian Legal Data Fair?

Sahil Girhepuje[1], Anmol Goel[2], Gokul S Krishnan[1], Shreya Goyal[3], Satyendra Pandey[1], Ponnurangam Kumaraguru[2] and Balaraman Ravindran[1]

[1]*IIT Madras*
[2]*IIIT Hyderabad*
[3]*American Express*

### Abstract
Recent advances and applications of language technology and artificial intelligence have enabled much success across multiple domains like law, medical and mental health. AI-based Language Models, like Judgement Prediction, have recently been proposed for the legal sector. However, these models are strife with encoded social biases picked up from the training data. While bias and fairness have been studied across NLP, most studies primarily locate themselves within a Western context. In this work, we present an initial investigation of fairness from the Indian perspective in the legal domain. We highlight the propagation of learnt algorithmic biases in the bail prediction task for models trained on Hindi legal documents. We evaluate the fairness gap using demographic parity and show that a decision tree model trained for the bail prediction task has an overall fairness disparity of 0.237 between input features associated with Hindus and Muslims. Additionally, we highlight the need for further research and studies in the avenues of fairness/bias in applying AI in the legal sector with a specific focus on the Indian context.

## 1. Introduction

Recent progress in language technology and NLP has enabled success in domains like law and justice. This has resulted in greater attention being paid to the development of curated, domain-specific datasets and language models for law resulting in the birth of a new subfield called *LegalNLP*. Language technologies in the law domain have far-reaching applications on all actors involved - judges, lawyers and citizens. Recent *LegalNLP* research has highlighted the impressive performance of assistive technologies on judgement prediction [1], prior case retrieval [2] and summarization [3]. Since legal outcomes have a significant impact on individuals, it is imperative to assess how *fair* and *biased* these technologies are. Deployment of legal technology without proper evaluation of bias and fairness can lead to unfair and biased outcomes as well as decreasing public trust in the legal system.

NLP systems trained on large corpora sourced from legal archives have a risk of learning historical social biases entrenched within the data and thus, perpetuating unfair decision-making in the future as well. It is already well-established that historically, legal data does not represent all social groups equally or fairly since the data reflects human and institutional biases pervasive in human society [4]. An evaluation and investigation of such biases is thus not only an essential step towards understanding historical social disparities but also mitigating any potential harms in the future.

This work presents the initial findings of our investigation into fairness in Indian legal data. We show evidence of algorithmic bias propagated in bail prediction by models trained on Hindi legal documents. We situate our methodology in a counterfactual fairness paradigm [5], wherein, we observe the change in model predictions on changing specific attributes in the input to the model corresponding to religious communities in India.

The next section, Section 2 includes a brief description of the related works. Section 3 describes the proposed methodology in detail while Section 4 includes the experimental results and discussion. Section 5 summarizes the paper with concluding remarks, limitations and scope for future avenues.

## 2. Previous Works

Several works based on language models and machine/deep learning have been put forth as applications in the legal domain in the last few years. Pretrained Transformer-based Large Language Models (LLMs) have recently been introduced for legal tasks [6, 7]. The LLMs are evaluated against several benchmark legal NLP tasks over Indian legal documents, namely, Legal Statute Identification from facts, Semantic segmentation of court judgments, and Court Judgement Prediction. Kapoor et al. [8] released HLDC: Hindi Legal Documents Corpus - a corpus of more than 900K Hindi legal documents that had been organized and sanitized. Summarization was used as a secondary task in conjunction with bail prediction as the primary task in multi-task learning. The publication also mentions various bail prediction task models that served as our starting point for preliminary analysis.

Bias and fairness aspects of machine learning in various domains have recently been explored. Ash et al. [9] recently highlighted the existence of in-group bias in the Indian district courts based on a causal analysis of over 5 million judicial records between 2010-2018. Kusner et al. [5] developed a Counterfactual Fairness framework for modeling fairness using tools from causal inference. They tried to capture fairness of a decision towards an individual in the actual world and a counterfactual world where the individual belonged to a different demographic group. On the same lines our work tries to explore the fairness of decision based on the religion of the accused. Bassett [10] do not examine bias in the straightforward context of claims of employment or housing discrimination but more fundamentally across every participant category within the legal system like clients, lawyers, judges, jurors, witnesses, and court personnel. Olaborede and Meintjes-Van der Walt [11] examine how cognitive heuristics affect judicial decision in South African courts-making with seven common manifestations of heuristics such as availability heuristics, confirmation bias, egocentric bias, anchoring, hindsight bias, framing, and representativeness. Lidén et al. [12] propose that confirmation bias may not only be present in the behaviours of individual agents in the judicial system but can also be recognized at a 'system-level' as an inability to self-correct, that is, an inability to acquit wrongfully convicted who appeal or petition for a new trial.

In this paper, we aim to investigate the bias involved in legal AI models in the Indian context

**Table 1**
Token Counts for the Original Text in the facts-and-arguments section. Each row represents the values for a given metric. The *Full Text* column contains aggregated values for the entire dataset. The *Bail granted only* and *Bail denied only* columns have values corresponding to the set of samples where bail was granted and denied respectively.

| Token Counts | Original Text | | |
|---|---|---|---|
| | Full Text | Bail granted only | Bail denied only |
| Mean | 344.7 | 395.9 | 315.5 |
| Median | 324 | 380.5 | 300 |
| Min | 4 | 4 | 5 |
| Max | 1233 | 1233 | 1193 |

for the task of bail prediction trained on Hindi case documents.

## 3. Methods

India is a country where there is significant diversity in multiple aspects such as religion, caste, language, ethnicity, etc. In such a scenario, a single AI model for the entire country will be pointless, and therefore, there is a crucial need for region-specific AI models. However, the diversity in the country also has an effect on data acquired, and therefore, trained AI models can result in skewing or bias. In this subsection, we perform a deeper analysis into the algorithmic biasing aspect of AI in legal setting in the Indian context. We particularly analyze the Religion axis of disparity and we motivated this idea in by making use of the Hindi Legal Documents Corpus (HLDC), a corpus introduced by Kapoor et al. [8]. HLDC is a corpus of more than 900K legal documents in Hindi which consists of documents that have been cleaned and structured to enable the development of downstream applications such as bail predictions of cases [8].

### 3.1. Dataset

We begin by sampling a random set of 10,000 cases from the corpus for our study. Some basic characteristics of the corpus are as shown in Table 1. We applied some basic NLP preprocessing techniques on the corpus and the characteristics of the corpus post preprocessing are as shown in Table 2. We perform our downstream tasks using two attributes: the *Facts and Arguments* section and the *Case Result*. The Facts contain the facts of the case and the defendant and prosecutor's arguments. We avoid the use of the *Judge's Summary* for the task of bail prediction since it may leak out information about the case result. The sampled dataset has 36% samples bail granted cases and 64% samples marked for bail denied cases. Hence we have a skewed distribution of labels.

### 3.2. LDA based Feature Engineering

To perform analyses of algorithmic bias in bail prediction model downstream task, we engineer a simple feature representation strategy that can enable interpretability in the later stages as well. We were motivated to represent the case using some keywords (crime description and personnel) present in the case text and also the category/type of the crime to indicate how critical the case is so as to decide whether the applicant is eligible for bail. These keywords can be seen in Table 3 .For extracting the category of crimes, we used a zero-shot classification using an existing transformer and for extracting the keywords from the case, we use the concept of topic modelling.

**Table 2**
Token Counts for the Preprocessed Text in the facts-and-arguments section. Preprocessing steps included removal of Hindi stopwords, punctuation marks, and filtering out weblinks using Regular Expressions.

| Token Counts | Preprocessed Text | | |
| --- | --- | --- | --- |
| | Full Text | Bail granted only | Bail denied only |
| Mean | 139.7 | 161.5 | 127.3 |
| Median | 127 | 149 | 117 |
| Min | 2 | 2 | 2 |
| Max | 759 | 661 | 759 |

Latent Dirichlet Allocation (LDA) is a topic modelling technique to extract latent topics in a document from a given text corpus using the idea that "a document is best described using a group of topics and topics are best described using a group of words". LDA is an unsupervised learning approach which requires the user to specify the number of topics to be extracted from the corpus. We applied LDA topic modelling to the 10,000 extracted cases from the HLDC corpus and analysed the topic clusters. For determining a near-optimal number of topics to be extracted, we varied the number of topics and analyzed the topic clusters using two metrics - the coherence score and perplexity scores. Coherence provides a measure of how interpretable that the topics are to humans, while perplexity measures how well a model predicts a sample [13]. We chose the number of clusters which gave the highest coherence score and therefore, a total of 85 topics are chosen for the corpus.

In contrast to hard-clustering algorithms like K-means, LDA follows a soft-clustering approach. Leveraging this fact, we extract the top 2 topics to which the cases belong to, i.e., each case sample is assigned a *dominant* and a *second-dominant topic.* We make use of "soft clustering" concept of a sample into multiple topics since two topics provide a richer understanding of a case than a single topic. The gensim [14] implementation of LDA also provides the "keywords" associated with each topic. We extracted the top 10 keywords that best describe a particular topic. For instance, if the keywords for a topic are – [मृतका (Mruthaka – deceased), दहेज (Dahej – dowry), शादी (Shaadi – marraige), पुत्री (Puthri – daughter), मांग (Maang – demand), हत्या (Hatya – murder), बहन (Behan – sister), मृत्यु (Mrityu – death), पत्नी (Patni – wife), ससुराल (Sasuraal – in law's house)], then it can be inferred that this topic includes cases about dowry-related deaths. Similarly, keywords such as [ग्राम (Gram), बरामद (Baramad – found) , नमूना (Namoona – sample), नशीला (Nasheela – intoxicating), पाउडर (Powder)] would belong to cases related to drugs. The top keywords from the dominant and second-dominant topics are then considered as categorical features for a case text, which is then used to train a classifier for the downstream task – bail prediction.

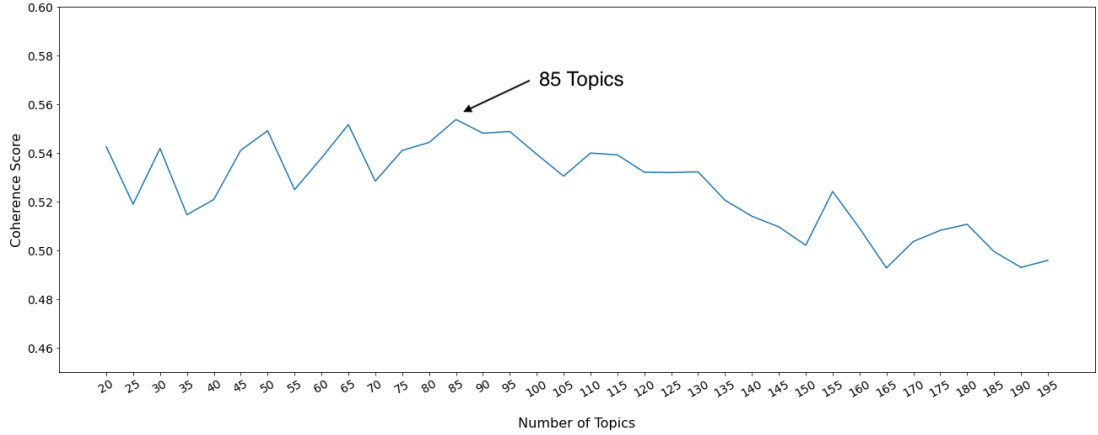

**Figure 1:** Choosing the Number of Topics by plotting the Coherence Score values against the number of topics given in LDA

For determining the category of crime, we assign labels to cases in a zero-shot manner using the HuggingFace implementation of the *xlm − roberta − large* model [15]. We referred to a standard categorization of crimes in India [16] and their corresponding translations were used as the target labels, such as दहेज (Dahej – dowry), दंगों (Danga – riot), चोरी (Chori – theft), हत्या (Hatya – murder), etc. Every case sample in the dataset is assigned a dominant, and a second-dominant case category/theme. To assign the category/theme of the crime for each case, we feed in the keywords from the topics obtained by LDA. This step is performed for both the dominant and second-dominant topics of a case and we have assigned two case categories/themes for each case.

As explained above, the LDA topic modelling helps us to extract the top keywords from the dominant topics. These keywords are words in the facts-and-arguments section of a case which could refer to various aspects of the case like the description of crimes, personnel/communities involved in the case and other entities. We extracted a total of five dominant keywords for each case - three belonging to the dominant topic and two from the second-dominant topic. Following this process, we can represent each case with seven categorical word features - five keywords and two case categories/themes. An illustration of the engineered feature representation is as shown in Figure 3. The first row indicates a case about a cheating/fraud case, whereas the second and third rows represent a theft and dowry/murder case respectively.

### 3.3. Training a Classifier

Since the Case Result of every sample in our dataset is known, we train a Decision Tree model for the task of bail prediction using the seven features stated above. Hyperparameters for the models are tuned using the Optuna library [17]. It is to be noted that we use Decision Trees for their high interpretabilty. In order to identify possible biases in the learned model - we train a model to obtain predictions, and also get predictions keeping all but one input features identical. We argue that - with all inputs same except the *Name*, if the model predicts a new label for the case, then the model strongly associates that Name with the change in prediction.

**Table 3**
Example of features representing a case, before encoding the Hindi tokens

| Keyword$_1$ | Keyword$_2$ | Keyword$_3$ | Keyword$_4$ | Keyword$_5$ | Theme$_1$ | Theme$_2$ | label |
|---|---|---|---|---|---|---|---|
| आधार (Aadhaar) | विरुद्ध (Viruddh) | दर्ज (Darj) | स्वीकार (Sweekaar) | None | धोखा (Dhokha) | वसूली (Vasooli) | 0 |
| दर्ज (Darj) | आधार (Aadhaar) | समर्थन (Samarthan) | तर्क (Tark) | आपराधिक (Aapradhik) | अपहरण (Apaharan) | दंगों (Dange) | 0 |
| मोहम्मद Mohammad | गौवध Gouwadh | पशुओं Pashuon | बरामद Baramad | बोर Bor | हत्या Hatya | दंगों Dange | 0 |

In other words, the model makes a change in prediction only because how it *perceives* that word. To test this hypothesis, we employ the following steps -

1. Identify a subset of cases from the dataset using the *theme*
2. Sample cases within this subset which have either a Hindu or a Muslim proper noun. Examples include किशोर (Kishore), कुलदीप (Kuldeep), जयराम (Jayaram), धीरेन्द्र (Dheerendra), नितिन (Nithin) (Hindu names) and अब्दुल (Abdul), अहमद (Ahmed), रिजवान (Rizwan), सलीम (Salim) (Muslim names)
3. For every case, we identify the
   a) True Label
   b) Model's Predicted Label
   c) Number of times the model's prediction changes when the proper noun is replaced with another *Hindu* proper noun
   d) Number of times the model's prediction changes when the proper noun is replaced with another *Muslim* proper noun

This exercise allows us to test whether the model has learnt to associate different characteristics with Hindu and Muslim proper nouns. We argue that if the model changes its predictions from 0 (bail dismissed) to 1 (bail granted) *more* for Muslim nouns replaced by Hindu nouns than Muslim nouns with Hindu nouns, then there exists a bias against Muslims. This bias may be inherent in the dataset, since the Decision Tree model learns to classify after being trained on the dataset itself.

### 3.4. Evaluating Fairness

We evaluate classifier fairness using *Demographic Parity* [cite]. Formally, Demographic Parity requires the outcome of a classifier to be independent of a protected attribute. The classifier should predict the same probability $P$ of outcome irrespective of sensitive attributes $A$.

$$P(\hat{Y}|A = 0) = P(\hat{Y}|A = 1) \tag{1}$$

Equation 1 mathematically defines Demographic Parity. Ideally, a *fair* classifier must predict the same outcome irrespective of demographic attributes in the input features. In this study, we use this formalization of disparity and use it as a metric to highlight and quantify algorithmic biases which might lead to unfair outcomes.

# 4. Results & Discussion

**Table 4**
Fairness Gap on Denial of Bail

| Crime | Fairness Gap |
|---|---|
| Murder | 0.071 |
| Theft | 0.48 |
| Dowry | 0.38 |
| Drugs | 0.054 |
| Overall | 0.237 |

For the experiments, we used only those cases where Hindu names or Muslim names appear in the keywords we extracted as part of the feature engineering process. For the extracted cases, we fed the features as input to the trained classifier and extracted the prediction probabilities for Bail Denied outcome. We then prepared a new set of inputs by changing the Hindu/Muslim names to Muslim/Hindu names and then calculated the probabilities for Bail Denied outcome from the classifier. For evaluating and quantifying the Fairness Gap, we calculate the differences of probabilities of bail prediction task outcomes for each sample in the test set, i.e, the difference of demographic disparities. The Fairness Gap reflects the deviation of the trained classifier away from ideal demographic parity (i.e., Fairness Gap must be zero in an ideal case). The results are analyzed based on the crimes/themes and are presented in Table 4. While it is good to note that for serious crimes like murder, the fairness gap shown by the model is minimal, other themes like dowry and theft themes show significant bias.

We also analyze the themes of हत्या (Hatya – murder) and दहेज (Dahej – dowry) further in depth by looking at the changes of prediction labels for various combinations of inputs with respect to changes in names from Hindu/Muslim community. The changes in labels for various combinations of names are presented in Tables 5 and 6.

We consider cases from theme *Hatya* - हत्या (Murder) where Hindu/Muslim names are present and for which the trained model predicted Bail Denied. Prediction changes for the selected cases are summarised in Table 5. The counts for prediction changes in the Hindu names column are higher than the counts of changes in the Muslim names column. This shows that for the cases where model predictions are 0, the model switches its predictions to 1 more often when the proper noun is replaced by a Hindu name instead of being replaced by a Muslim name, indicating that Hindus are more likely to get a bail granted.

Table 6 show prediction changes for selected cases from the theme दहेज (Dahej) - *Dahej* (Dowry), where the predictions from the model were Bail Granted. The counts for prediction changes in the Hindu names column are higher than the counts of changes in the Muslim names column. This shows that for the cases where model predictions are 1, the model switches its predictions to 0 more often when the proper noun is replaced by a Hindu name instead of being replaced by a Muslim name. Therefore, the model associates the negative aspect of *Dahej* more with Hindus as opposed to Muslims, indicating a learned bias against the Hindus.

**Table 5**
Changes in Predictions for Theme: *Hatya* (Murder)

| Predicted Label | Changed Label | Number of times model changes predictions when names replaced by | |
|---|---|---|---|
| | | Hindu names | Muslim names |
| 0 | 1 | 13 | 9 |
| 0 | 1 | 4 | 3 |
| 0 | 1 | 9 | 7 |
| 0 | 1 | 4 | 3 |
| 0 | 1 | 4 | 3 |
| 0 | 1 | 1 | 3 |
| 0 | 1 | 4 | 3 |
| 0 | 1 | 7 | 4 |
| 0 | 1 | 1 | 3 |
| 0 | 1 | 9 | 6 |
| 0 | 1 | 13 | 9 |

Labels – 0: Bail Denied; 1: Bail Granted

**Table 6**
Changes in Predictions for Theme: *Dahej* (Dowry)

| Predicted Label | Changed Label | Number of times model changes predictions when names replaced by | |
|---|---|---|---|
| | | Hindu names | Muslim names |
| 1 | 0 | 7 | 3 |
| 1 | 0 | 7 | 5 |
| 1 | 0 | 8 | 5 |

Labels – 0: Bail Denied; 1: Bail Granted

## 4.1. Ethical Considerations

It should be noted these results in no way indicate a bias in the judicial system of India. As we described in previous sections, the HLDC dataset contains data about Uttar Pradesh Court cases. It is a fact that Hindus are majority in population (especially in the state of Uttar Pradesh) and this could mean that there are larger number of cases among Hindus for the classifier to learn from, causing it to skew and show resulting in an algorithmic bias towards them. This strongly indicates that there is significant need for research to be conducted in the domain of Legal AI in Indian context. Fairness/bias studies need to be conducted exclusively for the Indian setting in not only developing models, but also in constructing datasets at both national and regional levels. Further, we also need to conduct research and analyses to identify sufficient de-biasing methods to apply for our Indian Legal AI models to ensure that the models in no way show any bias towards any groups of citizens, as these models have the potential to offer support in decision making for legal personnel and judiciary sector.

## 5. Conclusion

In this work, we presented our initial investigation into bias and fairness for Indian legal data and highlight preferentially encoded stereotypes that models might pick up in downstream tasks like bail prediction. We acknowledge that this does not indicate a bias in the judicial system at large and is only meant to be an evaluation of how algorithms may learn and propagate biases in their predictions. We leave an in-depth evaluation of fairness and bias in Indian legal data and models informed and grounded in legal domain expertise for future work.

We highlight the need for ethical considerations and research required in the direction of studying Indian legal data and models. More in depth analyses may be needed to understand fairness and bias issues in the AI models in legal sector focused on Indian scenario. Similarly, we also stress the importance of the need for algorithmic approaches to mitigate the bias learned by these models to ensure AI-enabled decision support in an effective and fair manner.

## Acknowledgements

This work was supported by iHub at IIIT Hyderabad, project O2-001.

## References

[1] V. Malik, R. Sanjay, S. K. Nigam, K. Ghosh, S. K. Guha, A. Bhattacharya, A. Modi, ILDC for CJPE: Indian legal documents corpus for court judgment prediction and explanation, in: Proceedings of the 59th Annual Meeting of the Association for Computational Linguistics and the 11th International Joint Conference on Natural Language Processing (Volume 1: Long Papers), Association for Computational Linguistics, Online, 2021, pp. 4046–4062. URL: https://aclanthology.org/2021.acl-long.313. doi:10.18653/v1/2021.acl-long.313.

[2] P. Jackson, K. Al-Kofahi, A. Tyrrell, A. Vachher, Information extraction from case law and retrieval of prior cases, Artificial Intelligence 150 (2003) 239–290.

[3] S. Klaus, R. Van Hecke, K. Djafari Naini, I. S. Altingovde, J. Bernabé-Moreno, E. Herrera-Viedma, Summarizing legal regulatory documents using transformers, in: Proceedings of the 45th International ACM SIGIR Conference on Research and Development in Information Retrieval, 2022, pp. 2426–2430.

[4] J. Sargent, M. Weber, Identifying biases in legal data: An algorithmic fairness perspective, arXiv preprint arXiv:2109.09946 (2021).

[5] M. J. Kusner, J. Loftus, C. Russell, R. Silva, Counterfactual fairness, Advances in neural information processing systems 30 (2017).

[6] I. Chalkidis, M. Fergadiotis, P. Malakasiotis, N. Aletras, I. Androutsopoulos, Legal-bert: The muppets straight out of law school, arXiv preprint arXiv:2010.02559 (2020).

[7] S. Paul, A. Mandal, P. Goyal, S. Ghosh, Pre-training transformers on indian legal text, arXiv preprint arXiv:2209.06049 (2022).

[8] A. Kapoor, M. Dhawan, A. Goel, A. T H, A. Bhatnagar, V. Agrawal, A. Agrawal, A. Bhattacharya, P. Kumaraguru, A. Modi, HLDC: Hindi legal documents corpus, in: Findings

of the Association for Computational Linguistics: ACL 2022, Association for Computational Linguistics, Dublin, Ireland, 2022, pp. 3521–3536. URL: https://aclanthology.org/2022.findings-acl.278. doi:10.18653/v1/2022.findings-acl.278.

[9] E. Ash, S. Asher, A. Bhowmick, S. Bhupatiraju, D. Chen, T. Devi, C. Goessmann, P. Novosad, B. Siddiqi, In-group bias in the indian judiciary (2022).

[10] D. L. Bassett, Deconstruct and superstruct: Examining bias across the legal system, UCDL Rev. 46 (2012) 1563.

[11] A. Olaborede, L. Meintjes-Van der Walt, Cognitive bias affecting decision-making in the legal process, Obiter 41 (2020) 806–830.

[12] M. Lidén, M. Gräns, P. Juslin, Self-correction of wrongful convictions: is there a 'system-level' confirmation bias in the swedish legal system's appeal procedure for criminal cases?—part i, Law, Probability and Risk 17 (2018) 311–336.

[13] D. Mimno, H. Wallach, E. Talley, M. Leenders, A. McCallum, Optimizing semantic coherence in topic models, in: Proceedings of the 2011 conference on empirical methods in natural language processing, 2011, pp. 262–272.

[14] R. Rehurek, P. Sojka, Gensim–python framework for vector space modelling, NLP Centre, Faculty of Informatics, Masaryk University, Brno, Czech Republic 3 (2011).

[15] A. Conneau, K. Khandelwal, N. Goyal, V. Chaudhary, G. Wenzek, F. Guzmán, E. Grave, M. Ott, L. Zettlemoyer, V. Stoyanov, Unsupervised cross-lingual representation learning at scale, CoRR abs/1911.02116 (2019). URL: http://arxiv.org/abs/1911.02116. arXiv:1911.02116.

[16] Wikipedia contributors, Crime in india — Wikipedia, the free encyclopedia, 2022. URL: https://en.wikipedia.org/w/index.php?title=Crime_in_India&oldid=1125259959, [Online; accessed 13-January-2023].

[17] T. Akiba, S. Sano, T. Yanase, T. Ohta, M. Koyama, Optuna: A next-generation hyperparameter optimization framework, in: Proceedings of the 25rd ACM SIGKDD International Conference on Knowledge Discovery and Data Mining, 2019.